\documentclass{article}
\usepackage{spconf,amsmath,graphicx,hyperref}
\usepackage{booktabs}
\usepackage{enumitem}   
\usepackage{colortbl}  %
\usepackage{xcolor}
\usepackage{multirow}
\usepackage{amssymb}
\usepackage{graphicx}
\usepackage{lipsum}
\usepackage{adjustbox}
\usepackage{tabularx}
\usepackage{array}
\usepackage{colortbl}
\usepackage{multirow}
\usepackage{capt-of}
\newcolumntype{C}{>{\centering\arraybackslash}X} 

\title{Cross Pseudo Labeling for Weakly Supervised Video Anomaly Detection}
\name{Dayeon Lee$^{*,1}$ \thanks{$^{*}$These authors contribute equally to this work.} \quad
   Donghyeong Kim$^{*,1}$ \quad
   Chaewon Park${^2}$\quad
   Sungmin Woo${^1}$\quad
   Sangyoun Lee${^1}$ \vspace{-0.2cm}}
\address{$^{1}$Department of Electrical and Electronic Engineering, Yonsei University, Seoul, Korea
\\$^{2}$Samsung Electronics, Suwon, Korea
\\{\tt\small $^{1}$\{rainbowgt7, 2donghyung87, smw3250, syleee\}@yonsei.ac.kr}
\\{\tt\small $^{2}$moogoopang@gmail.com}
}

\begin{document}
%
\maketitle
\begin{abstract}
Weakly supervised video anomaly detection aims to detect anomalies and identify abnormal categories with only video-level labels. We propose CPL-VAD, a dual-branch framework with cross pseudo labeling. The binary anomaly detection branch focuses on snippet-level anomaly localization, while the category classification branch leverages vision–language alignment to recognize abnormal event categories. By exchanging pseudo labels, the two branches transfer complementary strengths, combining temporal precision with semantic discrimination. Experiments on XD-Violence and UCF-Crime demonstrate that CPL-VAD achieves state-of-the-art performance in both anomaly detection and abnormal category classification.

\end{abstract}
\vspace{-0.5mm}
\begin{keywords}
 Vision language model, Video anomaly detection, Cross pseudo labeling
\end{keywords}
\vspace{-1.5mm}

\vspace{-3mm}
\section{Introduction}
\label{sec:intro}
\vspace{-3.5mm}

\indent~~ Video anomaly detection (VAD) aims to identify frames in long, untrimmed surveillance videos that contain dangerous or abnormal events. It has wide applications, including the detection of hazardous situations, unsafe behaviors, and inappropriate content in videos. However, training VAD models in a fully supervised manner requires dense frame-level annotations, which are extremely expensive and time-consuming to obtain at scale. To address this limitation, the field has recently shifted toward weakly supervised video anomaly detection (WSVAD) ~\cite{batchnorm_wsvad, vadclip,gcn,mist,mslnet,exploiting,tpwng}, which relies solely on video-level labels for training. 

Recent studies have extended WSVAD beyond simply distinguishing normal and abnormal frames, aiming to perform both binary anomaly detection and abnormal category classification at the same time. This shift reflects the need to address anomalies at different levels of granularity. Coarse-grained WSVAD focuses on deciding whether a segment is normal or abnormal, while fine-grained WSVAD goes further to identify the specific category of the abnormal event. VadCLIP is the first work to address both tasks simultaneously by leveraging CLIP, a vision–language model that encodes general semantic knowledge. VadCLIP adopts a dual-branch design to solve both tasks: a binary classification branch that performs traditional anomaly detection using video features for precise localization, and a category classification branch that leverages vision–language alignment to classify the specific type of abnormal event (e.g., fighting, abuse, shooting). 


\begin{figure}
    \centering
    \includegraphics[width=0.9\linewidth]{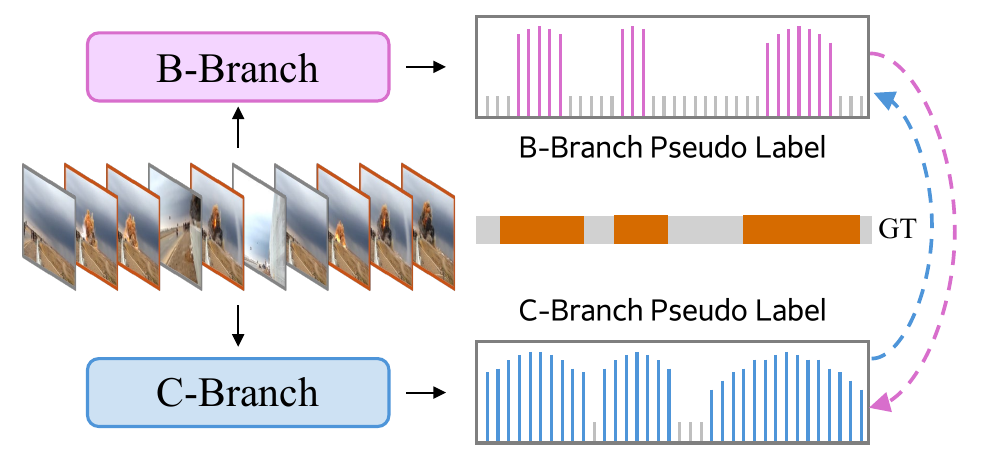}
    \vspace{-5.5mm}
    \caption{Cross pseudo labeling framework.}
    \vspace{-5.5mm}
    \label{fig_intro}
\end{figure}

\begin{figure*}[t]
\begin{center}
    \includegraphics[width=0.9\linewidth]{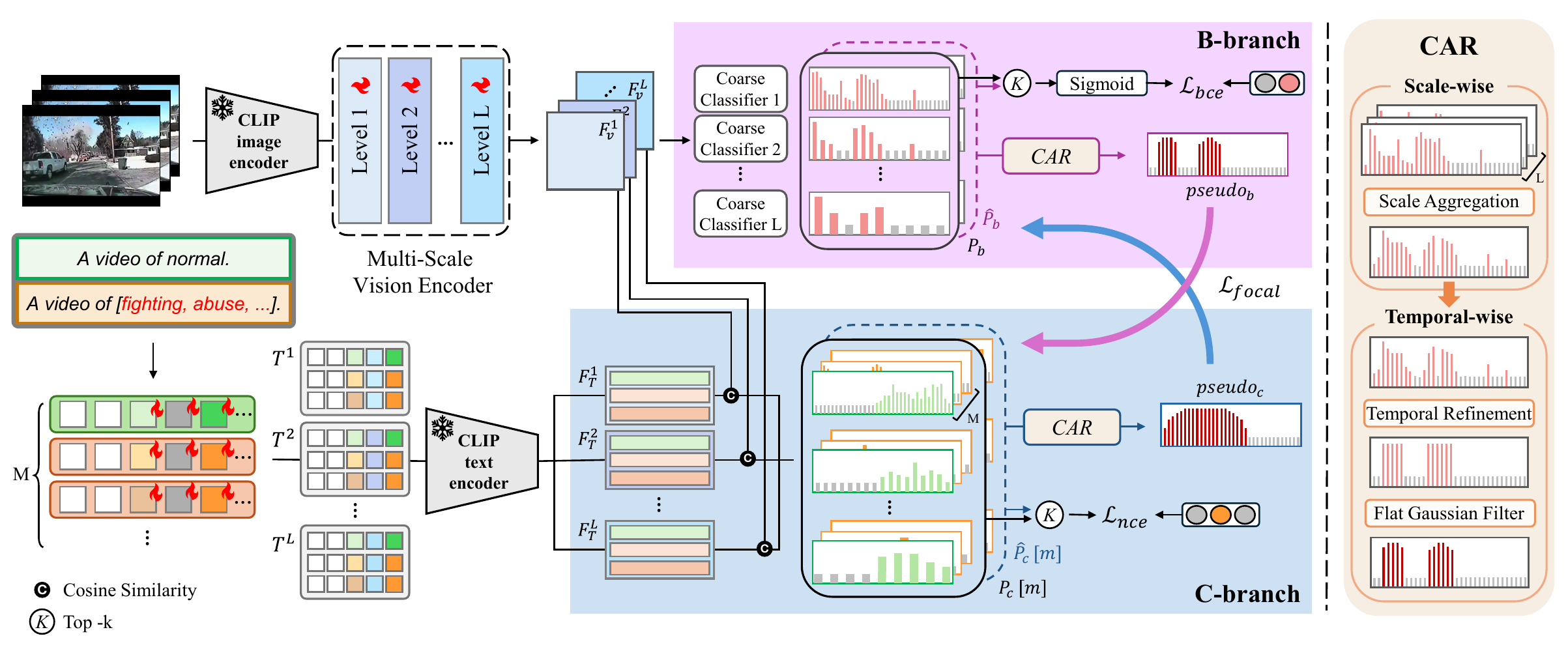}
\end{center}
\vspace{-7mm}
\caption{The framework of CPL-VAD and Consistency-Aware Refinement module.}
\vspace{-5mm}
\label{fig2}
\end{figure*}

In this work, we propose CPL-VAD, a dual-branch framework that enables the two branches to learn from each other through cross pseudo labeling. While the branches naturally complement one another, each also has inherent limitations.  The binary anomaly detection branch (B-branch) relies solely on video features, making it highly sensitive to temporal changes in motion and appearance. This enables high-precision anomaly localization, but it may miss subtle or semantically complex events, leading to false negatives. In contrast, the anomaly category classification branch (C-branch) leverages the CLIP text encoder and its broad semantic knowledge learned from large-scale image–text pairs. This enables recognition of diverse abnormal categories beyond the training data. However, since it focuses on classification without explicitly modeling temporal boundaries, it often overestimates anomaly regions and mislabels background as abnormal, resulting in false positives.

To overcome these limitations, CPL-VAD introduces cross supervision as shown in Fig.~\ref{fig_intro}, where each branch is trained not only with video-level labels but also with snippet-level pseudo labels generated by the other. The C-branch distills richer semantic cues into the B-branch, while the B-branch transfers the temporal precision to the C-branch. By exchanging the distinct characteristics of each branch, Cross pseudo labeling enables the B-branch to more reliably distinguish normal from abnormal events, and the C-branch to achieve more accurate temporal localization.

In addition, directly using snippet-level raw predictions of each branch as pseudo labels can result in instability, since they may contain outliers or inconsistencies across time. To alleviate this issue, we propose a Consistency-Aware Refinement (CAR) module that improves pseudo label quality by enforcing temporal consistency and filtering unreliable predictions. By providing cleaner and more reliable supervisory signals, this module stabilizes training and allows both branches to capture reliable video context.


\vspace{-4mm}
\section{Proposed Method}
\subsection{Overview}
\label{sec:overview}
\vspace{-2mm}

Fig.~\ref{fig2} illustrates the overall architecture of CPL-VAD. CPL-VAD consists of a B-branch for binary anomaly detection and a C-branch for classifying abnormal event categories. In addition, the framework adopts a two-stage design: the pseudo label generation stage, and the final prediction stage.

In both stages we use the same architecture. We extract snippet-level features $X_{\text{clip}} \in \mathbb{R}^{n \times d}$ using the frozen CLIP image encoder, where $n$ is the number of snippets and $d$ is the feature dimension. These features are fed into an ActionFormer~\cite{actionformer} based temporal encoder to obtain multi-scale video representations $\{F_v^i\}_{i=1}^{L}$, with $F_v^i \in \mathbb{R}^{t_i \times d}$ capturing temporal context at level $i$. The $i$ denotes the temporal scale: lower levels preserve finer temporal resolution with short-range dependencies, while higher levels capture coarser representations with longer temporal context. $L$ is the total number of levels in the multi-scale video encoder and $t_i$ is the temporal length at the corresponding scale. The $F^i_v$ are subsequently fed into a dual-branch architecture.

\noindent\textbf{Binary classification branch.}~ The B-branch applies MLP classifiers to each $F_v^i$, estimating multi-scale anomaly scores  $\hat{P}_{b} = \{{\hat p_b^{i}}\}_{i=1}^{L}$ and $P_b = \{{p_b^{i}}\}_{i=1}^{L}$ for the video features at each level. While $\hat{P_b}$ is used solely for generating pseudo labels, $P_b$ is the final predicted score of CPL-VAD. This branch is mainly responsible for snippet-level anomaly detection.

\noindent\textbf{Category classification branch.}~~The C-branch performs abnormal category classification via image-text alignment. However, since CLIP is not trained specifically for video anomaly detection, a direct application may fail to capture task-specific semantics. To address this, we use multi-level prompt learning, which maintains consistency across text prompts by leveraging shared tokens. We start from predefined templates such as “A video of [class],” where \textit{class} denotes categories like \textit{normal}, \textit{fighting}, or \textit{shooting}. These sentences are then transformed into scale-specific text prompts $T^i $, which is structured as follows:
\begin{center}
\vspace{-2mm}
\texttt{$T^i=$[A][video][of][class][$N_P, A_P$][$Q^i_P$]}.
\vspace{-2mm}
\end{center} 
Each prompt includes shared normal and abnormal tokens ($N_P$ and $A_P$) across all classes and levels, as well as a level-specific token $Q_P^i$ shared across classes at level $i$. By leveraging shared prompt tokens, we enable each text prompt to encode information about normality, abnormality, and temporal context. These prompts are passed through the frozen CLIP text encoder to yield textual features $F^i_T \in \mathbb{R}^{M \times d}$. $M$ denotes the number of categories. In the C-branch, we measure the similarity score between $F^i_T$ and $F^i_v$, which is the multi-class anomaly category score $\hat{P}_{c}[m] = \{{\hat p_c^{i}[m]}\}_{i=1}^{L}$ and ${P}_{c}[m] = \{{p_c^{i}[m]}\}_{i=1}^{L}$. Similar to the B-branch, $\hat{P}_{c}[m]$ is utilized for pseudo label generation, while ${P}_{c}[m]$ is adopted as the final prediction.

\vspace{-5mm}
\subsection{Cross Pseudo Labeling}\label{subsec:cross-pseudo} 
\vspace{-2mm}

We adopt a cross pseudo labeling strategy that enables complementary interaction between the B-branch and the C-branch. Since the effectiveness of such cross supervision fundamentally depends on the quality of the pseudo labels, we also propose a CAR module, which generates reliable pseudo labels, alleviating the inherent inconsistencies of weakly supervised predictions. Pseudo labels provide essential snippet-level guidance in weakly supervised settings, but they are not always reliable. In practice, they may include inconsistent boundaries across temporal scales, fragmented intervals or short spurious spikes.

\noindent \textbf{Cross pseudo label logits.} ~ We utilize $\hat{P}_{b}$ and $\hat{P}_{c}[m]$ trained solely with video-level ground truth, to generate snippet-level pseudo labels for each branch. Both logits are normalized and temporally upsampled to a fixed sequence length for alignment. Since C-branch generates multi-class predictions, the scores of all abnormal classes are summed to yield a single abnormality logit after softmax. The processed logits from each branch are independently fed into the CAR module. Specifically, $\text{\textit{pseudo}}_{c}$ is generated from $\hat{P}_{c}[m]$, while $\text{\textit{pseudo}}_{b}$ is generated from $\hat{P}_{b}$, both through the CAR module.

\noindent \textbf{Consistency-aware refinement.} 
To address inconsistency problem, the CAR module refines raw multi-scale logits into clear and reliable pseudo labels through two progressive steps: scale-wise aggregation and temporal refinement, as illustrated in Fig.~\ref{fig2}.

In the first step, predictions from different temporal scales are combined using an adaptive Gaussian RBF kernel that measures the similarity of scores across scales at each frame. Fine-scale predictions capture short-term details but are sensitive to noise, while coarse-scale predictions capture long-term context but may miss small events. Direct averaging can increase the influence of unreliable scales. To address this, the RBF kernel measures the similarity across scales, and its bandwidth is adaptively determined using the median absolute deviation. This adaptive setting enables the kernel to appropriately respond to the variation of scores across scales at each snippet. Each scale is then assigned a normalized weight based on its similarity with the others:
\vspace{-3mm}
\begin{equation}
w_t(i) = \frac{\sum_{j=0}^{L} e^{-\tfrac{(x_i - x_j)^2}{2\sigma^2}}}
               {\sum_{i=0}^{L}\sum_{j=0}^{L} e^{-\tfrac{(x_i - x_j)^2}{2\sigma^2}}},\; i \neq j ,
\vspace{-2mm}
\end{equation}
where $x_i$ and $x_j$ represent prediction scores from scales $i$ and $j$ at the same  snnipet $t$. The refined score is obtained by a weighted summation of multi-scale logits, which emphasizes consistent scales and suppresses outliers, yielding more robust aggregated logit.

Even with scale consistency, refined scores may still show temporal inconsistency such as short spikes or fragmented intervals. Such artifacts are common across many tasks that rely on pseudo labels~\cite{mist,exploiting,pseudo1,pivotal}, where labels generated under weak supervision often suffer from incomplete boundaries or unstable short segments. To address this, we adopt a temporal refinement strategy~\cite{pivotal} based on grouping and filtering, where consecutive abnormal segments are merged to ensure temporal continuity, and isolated short segments are removed. To further stabilize the predictions, flat Gaussian filter is applied at the boundaries of abnormal regions. This reduces abrupt score changes so that the center of abnormal segments provide strong supervision, while the boundary regions give weaker supervision due to their inherent uncertainty.

\vspace{-4mm}
\subsection{Cross branch Optimization}
\vspace{-2mm}
Both the pseudo label generation phase and the final prediction phase are trained under video-level supervision. We employ a similar loss function for both models, following prior work~\cite{vadclip}. In the B-branch, we select the most confident top-$K$ snippets from each prediction $P_b$ and $\hat{P}_{b}$. The binary cross-entropy loss with video-level annotations is then computed, denoted as $\mathcal{L}_{bce}$. In the C-branch, we adopt the multiple instance learning align mechanism proposed in previous work~\cite{vadclip}. For each class, we select the top-$K$ snippets with the highest confidence from $P_c[m]$ and $\hat{P}_{c}[m]$. We then compute a cross-entropy loss, denoted as $\mathcal{L}_{nce}$.

Unlike $\hat{P}_{b}$ and $\hat{P}_{c}[m]$, ${P}_{b}$ and ${P}_{c}[m]$ are optimized with snippet-level pseudo labels in addition to video-level labels. Each branch leverages the pseudo-label generated by the other. Specifically, $\textit{pseudo}_{b}$ are used to supervise the predictions of the C-branch ${P}_{c}[m]$, while $\text{\textit{pseudo}}_{c}$ guide the learning of the B-branch ${P}_{b}$. Since abnormal snippets are fewer than normal ones, we employ focal loss~\cite{focal} $\mathcal{L}_{focal}$ to mitigate class imbalance. For multi-class score $P_c[m]$, we take the abnormal confidence as $1-P_c\left[0\right]$ (i.e., the complement of the normal score) when computing the focal loss with $\text{pseudo}_b$. The overall loss of CPL-VAD is therefore formulated as:
\vspace{-1mm}
\begin{equation}
    \mathcal{L}_{total} = \mathcal{L}_{bce}+ \mathcal{L}_{nce} + \mathcal{L}_{focal} .
\end{equation}

\vspace{-0.9cm}
\subsection{Inference} 
\vspace{-2mm}
For inference, we employ the predictions $P_{b}$ for binary classification and $P_{c}[m]$ for multi-class classification. All multi-scale logits from each branch are upsampled to a fixed temporal length. They are then aggregated along the temporal axis, and averaged to obtain the unified logits $S_{ab}\in\mathbf{R}^{T}$ and $S_{cls}[m]\in\mathbf{R}^{T}$. The resulting $S_{ab}$ and $S_{cls}$ logits of all classes are subsequently passed through a sigmoid and a softmax function, respectively, to produce the normalized frame-level output scores.
\vspace{-2mm}
\begin{equation}
\vspace{-4mm}
S_{ab} = \frac{1}{L}\sum_{i=0}^{L}Up(p_{b}^{i}), \quad
S_{cls}\left[m\right] = \frac{1}{L}\sum_{i=0}^{L}Up(p_{c}^{i}\left[m\right])).
 \label{eq:logit1}
 \end{equation}
\vspace{-5mm}
\section{Experiments}
\begin{figure}
    \centering
    \includegraphics[width=\linewidth,height=0.6\linewidth,keepaspectratio=false]{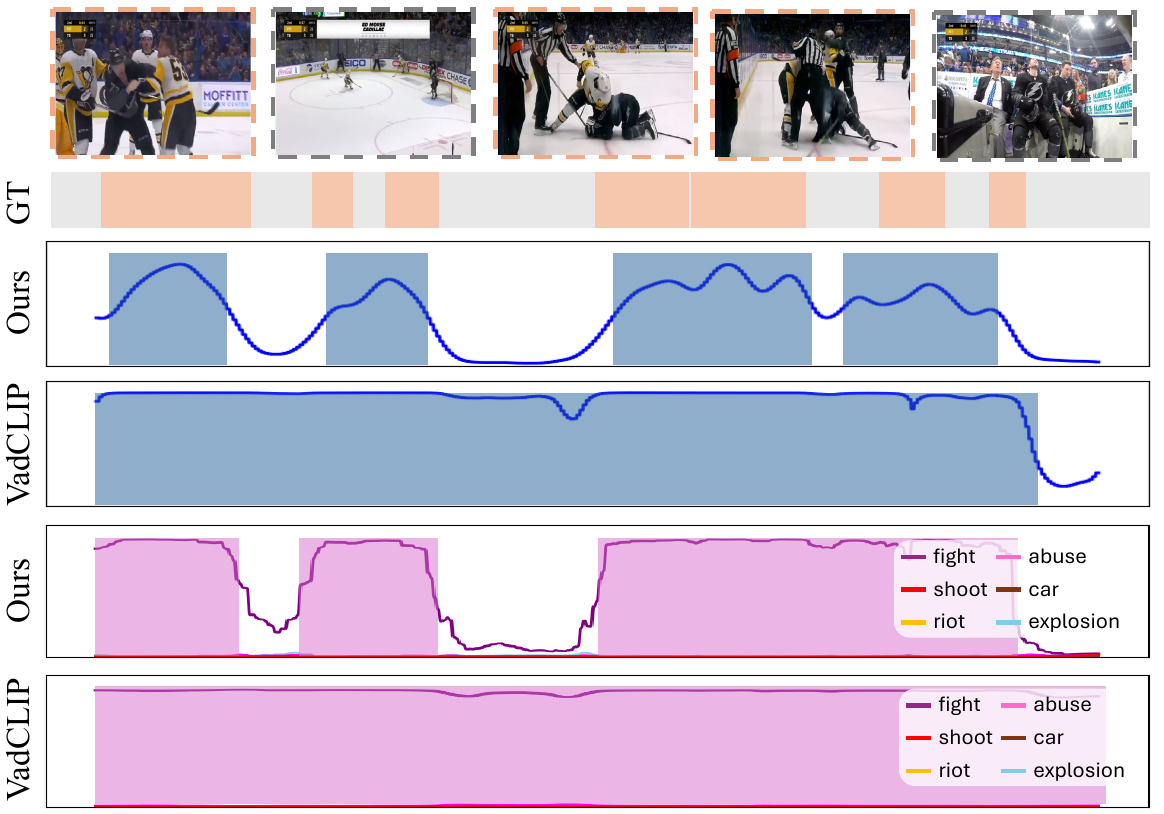}
    \vspace{-6mm}
    \caption{Qualitative results of CPL-VAD on XD-Violence.}
    \vspace{-2mm}
    \label{fig3}
\end{figure}
\vspace{-3mm}
\subsection{Implementation Details}
\vspace{-1mm}
\noindent \textbf{Datasets and metrics.}~~We evaluate CPL-VAD on two widely used benchmarks, XD-Violence~\cite{xd-violence} and UCF-Crime~\cite{ucf-crime}, where only video-level annotations are provided during training. For coarse-grained WSVAD, we use frame-level average precision (AP) on XD-Violence, and frame-level area under the curve (AUC) on UCF-Crime. For fine-grained WSVAD, we adopt the mean average precision (mAP) following the previous studies~\cite{vadclip, rad, avvd}.

\noindent \textbf{Training details.}~~We employed both text and video features extracted from pretrained CLIP (ViT-B/16). The number of multi-level layers was set to $L=6$. The number of snippet length is $n = 192$. The model was trained using Adam optimizer~\cite{kingma2014adam} with a learning rate of 1e-4. Training was conducted for $20$ epochs with a batch size of $32$.

\newcolumntype{L}[1]{>{\raggedright\arraybackslash\hsize=#1\hsize}X}
\newcolumntype{C}[1]{>{\centering\arraybackslash\hsize=#1\hsize}X}

\begin{table}[h]
\centering
\footnotesize{}
\begin{tabularx}{\linewidth}{C{1}|C{1}|C{1}}
\hline
\rowcolor[HTML]{EFEFEF}
\multicolumn{1}{c|}{\cellcolor[HTML]{EFEFEF}} &
\multicolumn{2}{c}{\cellcolor[HTML]{EFEFEF}\textbf{Dataset@Metric}} \\ \cline{2-3}
\rowcolor[HTML]{EFEFEF}
\multicolumn{1}{c|}{\cellcolor[HTML]{EFEFEF}Method} &
XD@AP(\%) & UCF@AUC(\%) \\ \hline
MSLNET~\cite{mslnet}   & 78.59 & 85.62 \\
Zhang et al.~\cite{exploiting}   & 78.74 & 86.22 \\
OVVAD~\cite{ovvad}   & 66.53 & 86.40 \\
CLIP-TSA~\cite{cliptsa} & 82.19 & 87.58 \\
IFS-VAD~\cite{Ifs}   & 83.14 & 86.57 \\
TPWNG~\cite{tpwng}   & 83.68 & 87.79 \\
VadCLIP~\cite{vadclip} & 84.51 & 88.02 \\ \hline
\textbf{CPL-VAD}     & \textbf{88.53} & \underline{88.24} \\ \hline
\end{tabularx}
\vspace{-4mm}
\caption{Coarse-grained comparisons on XD-Violence and UCF-Crime.}
\vspace{-3mm}
\label{tab:coarsexducf}
\end{table}

\begin{table}[hbt!]
\centering
\footnotesize
\resizebox{\linewidth}{!}{
\begin{tabular}{c|c|cccccc}
\hline
\rowcolor[HTML]{EFEFEF} 
&  & \multicolumn{6}{c}{mAP@IoU} \\ \cline{3-8}
\rowcolor[HTML]{EFEFEF} 
&   Method     & 0.1 & 0.2 & 0.3 & 0.4 & 0.5 & Avg \\ \hline
\cellcolor[HTML]{EFEFEF}  & RAD~\cite{rad}             & 22.72 & 15.57 &  9.98 &  6.20 &  3.78 & 11.65 \\
\cellcolor[HTML]{EFEFEF}   & AVVD~\cite{avvd}            & 30.51 & 25.75 & 20.18 & 14.83 &  9.79 & 20.21 \\
\cellcolor[HTML]{EFEFEF}XD    & VadCLIP~\cite{vadclip}  & 37.03 & 30.84 & 23.38 & 17.90 & 14.31 & 24.70\\
\cline{2-8} 
\cellcolor[HTML]{EFEFEF}    & \textbf{CPL-VAD}                  & \textbf{45.89} & \textbf{40.28} & \textbf{33.49} & \textbf{27.59} & \textbf{20.40} & \textbf{33.53} \\
\hline
\cellcolor[HTML]{EFEFEF} & RAD~\cite{rad}             &  5.73 &  4.41 &  2.69 &  1.93 &  1.44 &  3.24 \\
\cellcolor[HTML]{EFEFEF}    & AVVD~\cite{avvd}            & 10.27 &  7.01 &  6.25 &  3.42 &  \textbf{3.29} &  6.05 \\
\cellcolor[HTML]{EFEFEF}UCF    & VadCLIP~\cite{vadclip}  & 11.72 &  7.83 &  6.40 &  4.53 &  2.93 &  6.68 \\
\cline{2-8} 
\cellcolor[HTML]{EFEFEF}    & \textbf{CPL-VAD}                  & \textbf{17.77} & \textbf{11.26} & \textbf{9.75} & \textbf{5.93} & 2.23 & \textbf{9.39} \\ \hline
\end{tabular}}
\vspace{-4mm}
\caption{Fine-grained comparisons on XD-Violence and UCF-Crime.}
\label{tab:finexducf}
\vspace{-4mm}
\end{table}

\vspace{-5mm}
\subsection{Comparison with State-of-the-art Methods}
\vspace{-1.5mm}
We compared CPL-VAD with existing WSVAD approaches. For fairness, the comparison was restricted to models that utilize CLIP features. Table.~\ref{tab:coarsexducf} shows the comparison results of coarse-grained WSVAD. On the XD and UCF dataset, our model achieves a remarkable state-of-the-art performance. In particular, it shows notable improvements over prior methods~\cite{gcn, mist, mslnet, exploiting, tpwng} that adopt self-training strategies based on the pseudo labels. As shown in Table.~\ref{tab:finexducf}, our method also establishes new state-of-the-art results in fine-grained WSVAD, surpassing all existing methods. To intuitively demonstrate our superiority in both the B-branch and C-branch, Fig.~\ref{fig3} compares our model with VadCLIP, showing that our approach achieves more precise localization performance.

\begin{table}[]
\centering
\footnotesize
\resizebox{\linewidth}{!}{%
\begin{tabular}{cccc|cc} 
\hline
\rowcolor[HTML]{EFEFEF} 
\multicolumn{4}{c|}{\cellcolor[HTML]{EFEFEF}Pseudo label} & \multicolumn{2}{c}{\cellcolor[HTML]{EFEFEF}Dataset} \\ \hline
\rowcolor[HTML]{EFEFEF} 
B → C & C → B & self & CAR & \multicolumn{1}{c|}{\cellcolor[HTML]{EFEFEF}XD@AP,\;mAP(\%)} & UCF@AUC,\;mAP(\%) \\ \hline
& & & & \multicolumn{1}{c|}{76.50, 29.97} & 86.60, 6.68 \\
& & $\checkmark$ & $\checkmark$ & \multicolumn{1}{c|}{83.70, 31.47} & 87.34, 7.43 \\
$\checkmark$ & & & $\checkmark$ & \multicolumn{1}{c|}{83.30, 32.43} & 87.04, 7.29 \\
& $\checkmark$ & & $\checkmark$ & \multicolumn{1}{c|}{88.28, 27.75} & 87.61, 6.79 \\
$\checkmark$ & $\checkmark$ & &  & \multicolumn{1}{c|}{78.91, 31.86 }  &     86.42, 8.30 \\
$\checkmark$ & $\checkmark$ & & $\checkmark$ & \multicolumn{1}{c|}{\textbf{88.53, 33.53}} & \textbf{88.24, 9.39}\\  \hline
\end{tabular}}
\vspace{-4mm}
\caption{Ablation studies of pseudo label structure and consistency-aware refinement module.}
\vspace{-4mm}
\label{tab:ablation}
\end{table}

\begin{figure}[h]
    \centering
    \includegraphics[width=0.8\columnwidth]{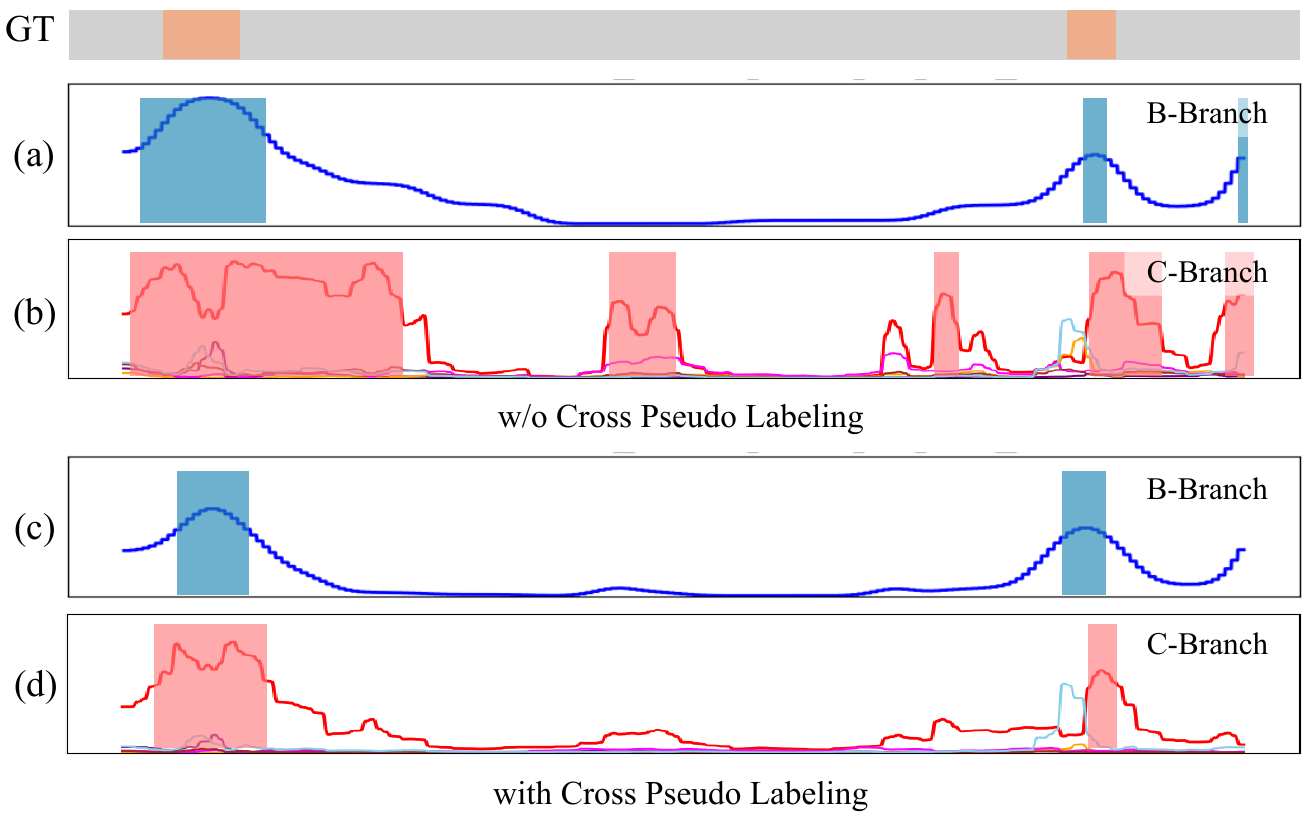}
    \vspace{-5mm}
    \captionof{figure}{Comparison of qualitative results without and with cross pseudo labeling.}
    \vspace{-6mm}
    \label{fig4}
\end{figure}
\vspace{-4mm}

\subsection{Ablation Studies}
\vspace{-1.5mm}
Table.~\ref{tab:ablation} presents the ablation results. The first row reports the pseudo label generation model as our baseline. We evaluate three variants of pseudo label usage: B→C (using pseudo labels from the B-branch to supervise the C-branch), C→B (using pseudo labels from the C-branch to supervise the B-branch), and self (each branch using its own pseudo labels). We also validate the effectiveness of CAR.

Both B→C and C→B lead to clear improvements by transferring complementary strengths between branches: B→C improves category classification, while C→B enhances binary anomaly detection. The bidirectional setting achieves the best overall performance. To verify that this gain is not simply from the presence of pseudo labels, we also test self-supervision. While self labels improve performance over the baseline, the gains are modest, suggesting that cross-branch supervision is essential for overcoming each branch’s inherent limitations. Additionally, we visualize qualitative prediction results to verify the effectiveness of cross pseudo labeling, as shown in Fig.~\ref{fig4}. Without cross pseudo labeling, each branch tends to under or over-estimate the anomaly segments. In contrast, with cross pseudo labeling, the B-branch discriminate between normal and abnormal events more robustly and the C-branch localizes events more precisely in time.

Finally, the CAR module consistently improves results across datasets and tasks. The improvement is especially pronounced on XD-Violence, which contains more numerous and fine-grained abnormal segments than UCF-Crime. In such cases, filtering out outliers and enforcing temporal consistency is crucial, and CAR substantially boosts pseudo label quality and overall detection accuracy.

\vspace{-5mm}
\section{Conclusion}
\vspace{-3mm}
In this paper, we presented CPL-VAD, a dual-branch framework for weakly supervised video anomaly detection. CPL-VAD utilizes cross pseudo labeling, enabling the B-branch and C-branch to complement each other. Furthermore, we introduced a consistency-aware refinement module that enhances both the reliability and temporal consistency of pseudo labels. 

\vfill\pagebreak
\small
\section{Acknowledgements}
This work was supported by Korea Planning \& Evaluation Institute of Industrial Technology(KEIT) grant funded by the Korea government(MOTIE) grant funded by the Korea government(MOTIE) (No. RS-2024-00442120, Development of AI technology capable of robustly recognizing abnormal and dangerous situations and behaviors during night and bad weather conditions) and the National Research Foundation of Korea (NRF) grant funded by the Korea government (MSIT)(No. RS-2024-00340745).

\bibliographystyle{IEEEbib}
\small
\bibliography{strings,refs}
\vspace{1em}

\end{document}